\documentclass{vgtc}                          %
\graphicspath{{figures/}{pictures/}{images/}{./}} 

\usepackage{times}                     
\usepackage{svg}
\usepackage{helvet} 

\usepackage{tabu}                      
\usepackage{booktabs}                  
\usepackage{lipsum}                    
\usepackage{mwe}                       

\usepackage{mathptmx}                  

\onlineid{0}

\vgtccategory{Research}

\vgtcinsertpkg

\title{A2P-Vis: an Analyzer-to-Presenter Agentic Pipeline for Visual Insights Generation and Reporting
}

\author{Shuyu Gan\thanks{e-mail: gan00067@umn.edu} %
\and Renxiang Wang\thanks{e-mail: renxiang428@gmail.com, This work was done during Renxiang’s
internship at University of Minnesota.} %
\and James Mooney\thanks{e-mail: moone174@umn.edu}
\and Dongyeop Kang\thanks{e-mail: dongyeop@umn.edu}}
\affiliation{\scriptsize University of Minnesota}

\teaser{
  \centering
  \includegraphics[width=\linewidth, trim=0 120 0 120,clip]{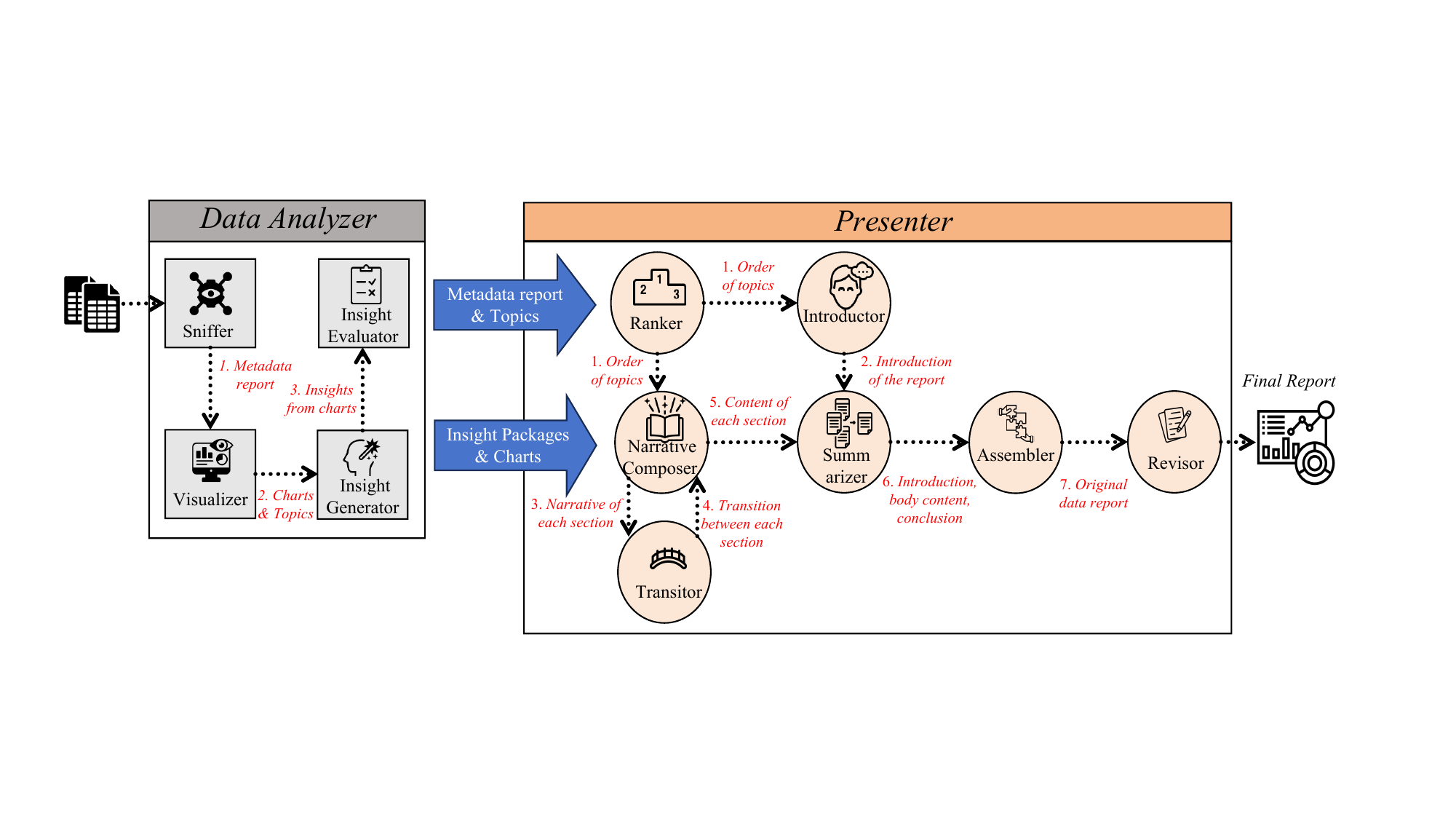}
  \caption{The workflow of \textit{\textbf{A2P-Vis}}.}
  \label{fig:teaser}
}

\abstract{
    Automating end-to-end data science pipeline with AI agents still stalls on two gaps: generating insightful, diverse visual evidence and assembling it into a coherent, professional report. We present \textbf{\textit{A2P-Vis}}, a two-part, multi-agent pipeline that turns raw datasets into a high-quality data-visualization report. The \textbf{Data Analyzer} orchestrates profiling, proposes diverse visualization directions, generates and executes plotting code, filters low-quality figures with a legibility checker, and elicits candidate insights that are automatically scored for depth, correctness, specificity, depth and actionability. The \textbf{Presenter} then orders topics, composes chart-grounded narratives from the top-ranked insights, writes justified transitions, and revises the document for clarity and consistency, yielding a coherent, publication-ready report.  Together, these agents convert raw data into curated materials (charts + vetted insights) and into a readable narrative without manual glue work. We claim that by coupling a quality-assured Analyzer with a narrative Presenter, A2P-Vis operationalizes co-analysis end-to-end, improving the real-world usefulness of automated data analysis for practitioners. For the complete dataset report, please see: \url{https://www.visagent.org/api/output/f2a3486d-2c3b-4825-98d4-5af25a819f56}.
} 


\begin{document}
\firstsection{Introduction}
\maketitle

LLM-based AI agents have recently been applied to automate data-centric workflows, from early-stage pipeline construction such as Google’s Data Science Agent~\cite{google_dsa_labs_2025} and related LLM-based systems~\cite{guo2024dsagent, zhang2023datacopilot, hong2024datainterpreter} to domain-specific applications in genomics~\cite{liu2024genotex}, machine learning benchmarking~\cite{liu2025hypobench} and healthcare~\cite{merrill2024wearable_llm_agents}. Beyond these pipelines, recent systems such as~\cite{wang2025chartinsighterapproachmitigatinghallucination, Zhao_2025_LEVA, Zhao_2025_LightVA} highlight the emerging role of LLMs in structuring, verifying, and contextualizing insights for visual analytics. Yet, two persistent gaps remain: (i) producing diverse, evidence-rich visualizations with non-trivial insights,
and (ii) assembling these materials into a coherent, professional report. \\
We introduce A2P-Vis, a two-part, multi-agent workflow designed to close both gaps, as described in Figure~\ref{fig:teaser}. Our design is
motivated by how data scientists carefully inspect initial seed data
and come up with final insights from trial and errors (pick the best
one among generated candidates). Our contributions are as follow:
\begin{enumerate}
    \item \textbf{Data Analyzer} explores possible visualization directions, generates candidate plots, filters low-quality charts, finds the best insight from candidates and evaluates insights.
    \item \textbf{Presenter} sequences topics, writes chart-grounded narratives with reasoned transitions, makes conclusions, and polishes the prose as a coherent visual story of findings.
\end{enumerate}

\section{System Overview}
Now we dive into each part of the framework.
\subsection{Data Analyzer}
\begin{figure}[t]
  \centering
  \includegraphics[width=\columnwidth]{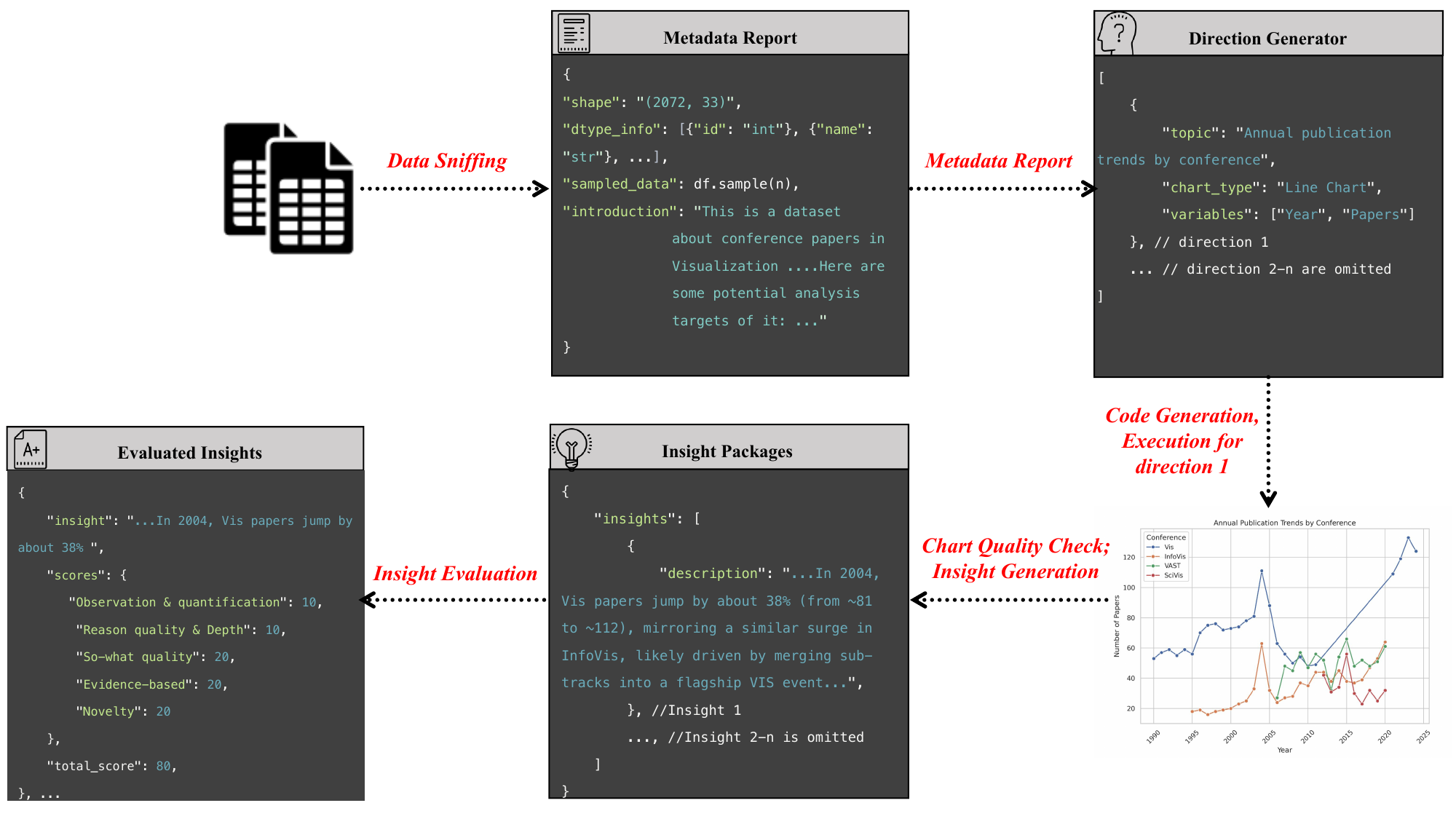}
  \caption{Data Analyzer in Action (From metadata profiling to graded insights).}
  \label{fig:analyzer-snapshot}
\end{figure}

This module ingests the dataset and outputs a metadata report, a diverse set of visualization topics, executable plots, and structured insights with scores. As shown in Figure~\ref{fig:analyzer-snapshot}, the process is demonstrated on a real dataset with grounded outputs at each stage. 

\noindent\textbf{Sniffer}
The \emph{Sniffer} performs lightweight dataset profiling to establish a reliable contract for downstream agents. It inspects the raw table to extract shape, column names and inferred types. From these basic features and randomly sampled data it produces a concise \emph{metadata report}, which introduces the dataset, enumerates its attributes, and outlines plausible analysis themes.
This report guides the \emph{Visualizer} to find visualize directions and use valid columns and sensible encodings for code generation, reducing hallucinations and preventing routine failures (e.g., empty plots, degenerate scales). Additionally, it addresses context and cost: rather than streaming the full dataset into the model, downstream components consume the compact profile, which is sufficient for planning visualizations and generating code without exposing raw records. In practice, the Sniffer’s output serves as a schema contract and quality gate, improving the robustness of code generation and keeping subsequent analyses aligned to a consistent data view.

\noindent\textbf{Visualizer}
Given the metadata report, the \emph{Visualizer} turns profile-level signals into reliable figures through a tightly coupled, four-step flow. First, the \emph{direction generator} proposes concrete analysis targets and topics, emitting machine-readable guidance for each direction (\texttt{topic}, \texttt{chart\_type}, \texttt{variables}). Next, the \emph{code generator} compiles that guidance into directly executable script. The script is then executed by the \emph{executor}, which logs outcomes and, upon failure. If errors are raised, it will invoke the \emph{rectifier} to repair the code strictly according to the error trace and then execute it again. Finally, each figure passes through a chart judger that validates quality one by one, checking for meaningless charts.

\noindent\textbf{Insight Generator \& Evaluator.}
Starting from each quality gated figure, the \emph{Insight Generator} produces 5–7 candidate insights using a \emph{three sentence} structure: (1) an observation with chart evidence and an approximate effect size; (2) a hedged, plausible reason anchored in chart context or domain context; and (3) a ``so what'' that delivers either a concrete next step, a short horizon prediction, or a precise implication. The \emph{Insight Evaluator} then scores each insight with an integer rubric aligned to this template across four criteria: \emph{Correctness \& Factuality}, \emph{Specificity \& Traceability}, \emph{Insightfulness \& Depth}, and \emph{``So what'' Quality}. It ranks candidates by total score and returns the top $3$ per chart as the final output.

\subsection{Presenter}
\begin{figure}[t]
  \centering
  \includegraphics[width=\columnwidth]{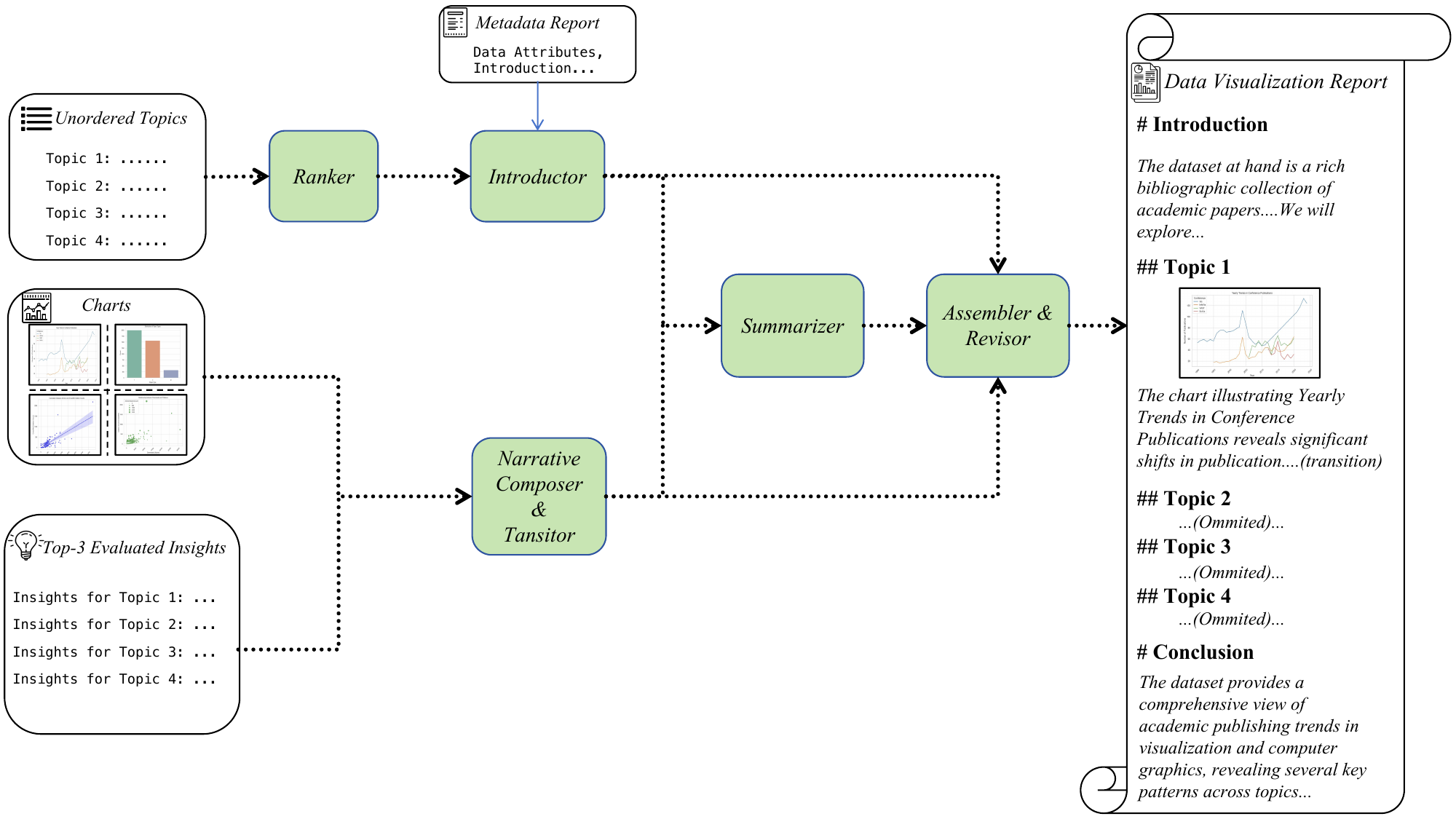}
  \caption{Presenter in Action(From ranked topics to a narrative report).}
  \label{fig:presenter-snapshot}
\end{figure}

This module takes the outputs of the \emph{Data Analyzer} (the metadata report, visualization topics, and quality-gated charts with scored insights) and assembles them into a publication-ready data visualization report. Following an "overview$\rightarrow$ sections$\rightarrow$ summary" sequence, it orders topics, drafts the introduction, writes chart-grounded sections with brief transitions, summarizes key takeaways, and finalizes the document with a revision pass. Figure~\ref{fig:presenter-snapshot} illustrates the workflow and the report layout.

\noindent\textbf{Ranker}
Given an unordered set of topics from the Data Analyzer, the \emph{Ranker} determines a coherent sequence for narration by analyzing relationships among topics, such as shared variables, temporal order, or thematic similarity.  
It outputs an ordered list of topics that provides a logical flow for the report, serving as the backbone for the \emph{Introductor} and subsequent modules.

\noindent\textbf{Introductor}
Given the ranked topic sequence and the metadata report, the \emph{Introductor} drafts the opening of the report and sets the roadmap. It first presents the dataset’s basic properties such as size, time span, and key attributes, grounded in the metadata. It then previews the analysis in the given order, offering one–two sentence teasers for each forthcoming topic (what it examines and why it matters). The result is a concise outline that bridges the global context to the individual topic sections and hands off naturally to the \emph{Narrative Composer}.

\noindent\textbf{Narrative Composer \& Transitor} The \emph{Narrative Composer} integrates the top three evaluator scored insights for each chart into a coherent, chart grounded subsection. It follows a claim to evidence to implication pattern, cites the figure explicitly, retains approximate magnitudes and axis references, and standardizes terms and units across sections. The \emph{Transitor} then adds one or two bridge sentences that connect the current section to the next by referencing shared variables, time windows, or meaningful contrasts, improving continuity without inventing links.

\noindent\textbf{Summarizer}
Building on the introduction and per-topic narratives, the \emph{Summarizer} produces an executive summary and closing conclusions: it consolidates key findings by topic, highlights cross-topic links or contrasts. The summary ties back to the preceding sections, echoing the main points to provide a cohesive close.

\noindent\textbf{Assembler} The \emph{Assembler} compiles the introduction, section content, and summary into a complete Markdown report. It applies a consistent heading hierarchy, embeds figures at their designated locations with captions. It also generates elements such as date, headers, footers, and page numbers to standardize the report format.

\noindent\textbf{Revisor}
Given the assembled Markdown draft, the \emph{Revisor} applies a multi-pass, chain-of-thought~\cite{wei2022chainofthought}
style revision, normalizing structure, smoothing transitions, polishing narrative and wording, and enforcing style consistency to produce the final report.

\section{Conclusion}
We introduce \emph{\textbf{A2P-Vis}}, an analyzer-to-presenter agentic pipeline for producing coherent, publication-ready data visualization reports. The \emph{Data Analyzer} profiles metadata, generates and executes visualization directions, and yields quality-gated charts with evaluator-scored insights; the \emph{Presenter} ranks topics, drafts the introduction, composes chart-grounded sections with light transitions, summarizes key takeaways, and finalizes the document via revision. This design emphasizes verifiable insight (structure + rubric), diversity in visualization directions and end-to-end reliability from raw data to narrative report. 
\acknowledgments{The authors wish to thank Qianwen Wang for discussions and valuable suggestions during development of this
work.}
\bibliographystyle{abbrv-doi}

\bibliography{template}
\end{document}